\PassOptionsToPackage{numbers,compress}{natbib} 
\documentclass{article}
 \usepackage[preprint]{neurips_2025}

\usepackage[utf8]{inputenc} 
\usepackage[T1]{fontenc}    
\usepackage{hyperref}       
\usepackage{url}            
\usepackage{booktabs}       
\usepackage{amsfonts}       
\usepackage{nicefrac}       
\usepackage{microtype}      
\usepackage{xcolor}         

\usepackage[utf8]{inputenc} 
\usepackage[T1]{fontenc}    
\usepackage{hyperref}       
\usepackage{url}            
\usepackage{booktabs}       
\usepackage{amsfonts}       
\usepackage{nicefrac}       
\usepackage{microtype}      
\usepackage{xcolor}         
\usepackage{graphicx} 
\usepackage{subcaption}  
\usepackage{amsmath}       
\usepackage{makecell}

\usepackage{graphicx}
\usepackage{booktabs}
\usepackage{multirow}

\title{STAR: Stage-Wise Attention-Guided Token Reduction for Efficient Large Vision-Language Models Inference}

\author{%
Yichen Guo, Hanze Li, Zonghao Zhang, Jinhao You, Kai Tang, Xiande Huang\thanks{Corresponding author. } \\
  De Artificial Intelligence Lab\\
  \texttt{\{yichenguo,lihanze,zonghaosevenzhang,jinhaoyou,kai,xdhuang\}@dail.email} \\
}

\begin{document}

\maketitle

\begin{abstract}

Although large vision–language models (LVLMs) leverage rich visual token representations to achieve strong performance on multimodal tasks, these tokens also introduce significant computational overhead during inference. Existing training-free token pruning methods typically adopt a single-stage strategy, focusing either on visual self-attention or visual-textual cross-attention. However, such localized perspectives often overlook the broader information flow across the model, leading to substantial performance degradation, especially under high pruning ratios. In this work, we propose \textbf{STAR} (\textbf{St}age-wise \textbf{A}ttention-guided token \textbf{R}eduction), a training-free, plug-and-play framework that approaches token pruning from a global perspective. Instead of pruning at a single point, STAR performs attention-guided reduction in two complementary stages: an early-stage pruning based on visual self-attention to remove redundant low-level features, and a later-stage pruning guided by cross-modal attention to discard task-irrelevant tokens. This holistic approach allows STAR to significantly reduce computational cost while better preserving task-critical information. Extensive experiments across multiple LVLM architectures and benchmarks show that STAR achieves strong acceleration capabilities, while maintaining, and even improving performances in various cases.

\end{abstract}

\section{Introduction}

With the rapid development of Large Language Models (LLMs) \cite{touvron2023llama,touvron2023llama2,grattafiori2024llama3,brown2020languagegpt3,achiam2023gpt4,team2023gemini,qwen2,qwen2.5,liu2024deepseek}, their outstanding reasoning and conversational abilities promote the advancement of Large Vision-Language Models (LVLMs). By integrating vision encoders (such as CLIP \cite{radford2021learning}) with LLMs, representative LVLMs such as LLaVA \cite{liu2023visual} and InstructBLIP \cite{instructblip} have been proposed. These LVLMs demonstrate remarkable performance on multimodal tasks, including image captioning and visual question answering(VQA) \cite{goyal2017makingVQAV2,gurari2018Vizwiz,hudson2019gqa,singh2019towardsTextVQA,lu2022learnScienceQA}. However, in contrast to textual tokens, visual tokens often exhibit significant redundancy, which not only increases computational complexity, but also hampers inference efficiency. 

To alleviate the computational burden, recent researches have focused on reducing the number of visual tokens by adaptively eliminating redundant tokens \cite{rao2021dynamicvit,liang2022notEviT,li2024llama-VID,alayrac2022flamingo,kong2022spvit}. However, these approaches typically introduce extra parameters or require fine-tuning, which may hinder their practicality in large-scale deployment. As an alternative, training-free methods have been proposed, employing attention-based ranking strategies to assess the importance of visual tokens and discard those with lower attention scores \cite{li2024tokenpacker,bolya2022tokenToMe,shang2024llava-prumerge,chen2024imagefastv,zhang2024clsFastervlm,zhang2024sparsevlm}. These techniques are applied either immediately after the vision encoder or progressively throughout the decoder layers. While effective in reducing computational complexity and inference latency, such approaches often compromise the model’s ability to capture comprehensive visual information. Specifically, aggressively pruning visual tokens before feeding them into the LLM can lead to substantial loss of visual information \cite{li2024inferencequecc,zhang2024clsFastervlm}. On the other hand, early-stage pruning within the decoder may hinder the model's ability to fully exploit visual cues via cross-modal attention \cite{chen2024imagefastv,zhang2024sparsevlm}.

\begin{figure}
  \centering
  \includegraphics[width=1\linewidth]{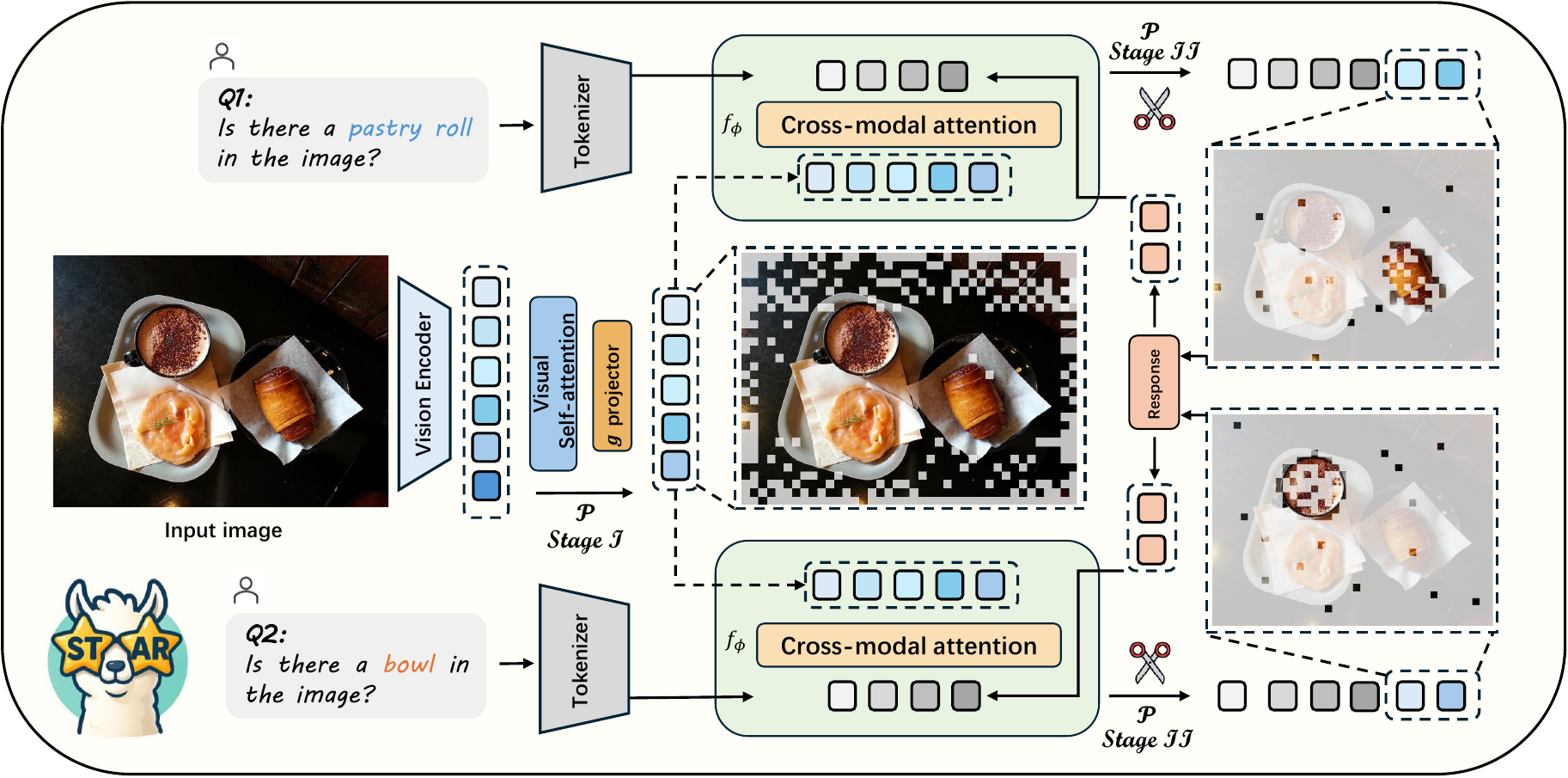}

  \caption{
\textbf{The framework of STAR.} The two-stage pruning includes visual self-attention pruning to remove redundant tokens, followed by cross-modal attention pruning to discard task-irrelevant ones.}
  \label{fig:overall_structure}
\vspace{-3mm}
\end{figure}

In this paper, we propose \textbf{STAR} (\textbf{St}age-wise \textbf{A}ttention-guided token \textbf{R}eduction), a novel framework for pruning redundant visual tokens in LVLMs, as shown in Figure~\ref{fig:overall_structure}. STAR adopts a holistic, two-stage attention-guided pruning strategy to efficiently reduce visual token redundancy. In the first stage, it performs conservative pruning based on visual self-attention, aiming to eliminate low-level redundant features. In the second stage, a more aggressive pruning is conducted at an intermediate decoder layer, where cross-modal attention helps identify and discard task-irrelevant tokens. This stage-wise design allows STAR to substantially reduce computational overhead while effectively retaining task-relevant visual information.

As a training-free and plug-and-play token pruning framework, STAR can be seamlessly integrated into a variety of LVLMs, including LLaVA-1.5 \cite{liu2023visual} and LLaVA-NEXT \cite{liu2024llavanext}. Extensive experiments on diverse vision–language benchmarks demonstrate that STAR consistently outperforms existing single-stage pruning methods that rely solely on visual self-attention or cross-modal attention, across a broad spectrum of token reduction ratios. For example, when applied to LLaVA-1.5-7B, STAR is able to prune up to 95\% of vision tokens, reducing inference FLOPs by more than \textbf{28.7}\% while retaining over \textbf{97.95}\% of the baseline performance across 7 different tasks. Furthermore, STAR surpasses prior methods under high reduction ratios, highlighting the effectiveness of stage-wise attention-guided reduction framework. Our contributions are summarized as follows:
\begin{enumerate}

    \item We identify critical limitations of current single-stage pruning approaches that rely solely on a single type of attention mechanism. Our analysis reveals that these methods often lead to substantial loss of essential visual information. We demonstrate that such degradation is not inevitable and can be effectively mitigated through a more precise, stage-aware pruning strategy.

    \item We propose STAR, a novel two-stage attention-guided pruning framework. STAR first performs conservative pruning based on visual self-attention after the vision encoder to eliminate low-importance tokens. It then applies a more aggressive pruning step at an intermediate decoder layer, leveraging visual-to-text cross-attention to discard task-irrelevant tokens while preserving critical information for downstream tasks.

    \item We evaluate STAR across multiple large vision-language models, including LLaVA-1.5 and LLaVA-NeXT, and validate its effectiveness on a range of benchmarks. Experimental results show that STAR consistently achieves substantial FLOPs reduction while maintaining strong performance across various token reduction ratios.

\end{enumerate}

\section{Related Work}

\subsection{Large Vision-Language Models (LVLMs)}

The remarkable progress of LLMs has fueled the emergence of large vision-language models (LVLMs). A typical LVLM architecture consists of a vision encoder\cite{dosovitskiy2020imageViT} that transforms input images into a sequence of visual tokens, and a large language model(LLM) \cite{touvron2023llama,vicuna2023} for text generation or comprehension. While processing input images in vision encoder, a multi-modal projector \cite{liu2023visual} is utilized to align visual tokens to the textual embedding space, enabling joint processing by the LLM. While this architecture enables LLMs to perceive visual representations, it also introduces long sequences of visual tokens, which leads to significant computational overhead. For instance, encoding a 336×336 image typically yields around 576 tokens \cite{liu2023visual}, and doubling the resolution can easily inflate the token count to several thousand \cite{liu2024llavanext}. Moreover, a large portion of image tokens are often redundant, leading to unnecessary computational cost and a higher risk of hallucinations \cite{fu2023mme,li2023evaluatingPOPE,yu2023mmvet,liu2024mmbench}. These issues make LVLMs inference more expensive and less reliable, highlighting the critical importance of optimizing the inference process for deploying LVLMs in resource-limited real-world scenarios \cite{goyal2017makingVQAV2,singh2019towardsTextVQA,lu2022learnScienceQA,hudson2019gqa,gurari2018Vizwiz}.

\subsection{Token Reduction for LVLMs}

Token reduction is a general framework that aims to alleviate computational overhead by reducing the number of visual tokens. Several studies \cite{wang2025dymu,koner2024lookupvit,chen2023diffrate,jie2024token} focus on visual token compression by encoding spatially structured image patches into more compact representations. For instance, TokenPacker \cite{li2024tokenpacker} arranges tokens hierarchically in a coarse-to-fine manner for token reduction, QueCC \cite{li2024inferencequecc} leverages a cross-attention mechanism incorporating query-based convolutional downsampling to compress tokens. Similarly, token merging \cite{cao2023pumer,shang2024llava-prumerge,wang2025dymu,bolya2022tokenToMe} aggregates redundant visual tokens based on attention scores or embedding similarity. For instance, ToMe \cite{bolya2022tokenToMe} employs a lightweight bipartite matching algorithm to merge tokens based on embedding similarity, significantly accelerating inference with minimal loss in accuracy. LLaVA-PruMerge \cite{shang2024llava-prumerge} introduces an adaptive token reduction approach that first prunes less informative visual tokens using attention sparsity and then merges the remaining tokens based on key similarity. 

Token pruning \cite{huang2024ivtp,dhouib2025pact,cao2024madtp,chen2024imagefastv,zhang2024clsFastervlm,zhang2024sparsevlm,xing2024pyramiddrop} selectively removes redundant visual tokens by assessing their importance without modifying the original information flow. FastV \cite{chen2024imagefastv} first reveals the redundancy phenomenon of visual tokens and removes visual tokens with low attention scores after the second layer of LLM. SparseVLM \cite{zhang2024sparsevlm} removes distractions from text prompts and utilizes text-visual attention to progressively sparsify visual tokens in LLM. FasterVLM \cite{zhang2024clsFastervlm} ranks visual tokens by the [CLS]-to-image attentions in the visual encoder and prunes low-salience tokens before they are passed to LLM. However, these pruning methods are overly aggressive and compromise model's performance for lower latency and computation overhead. In this work, we focus on optimizing LVLMs inference with Stage-wise image token pruning without sacrificing model's capability.

\section{Motivation}

\subsection{Preliminaries}
Large Vision-Language Models (LVLMs) typically consist of two primary components: a visual encoder \cite{radford2021learning} and a decoder-only large language model (LLM) \cite{vicuna2023}. While both modules rely on transformer-based architectures \cite{vaswani2017attentionsisallyouneed}, they differ subtly in the application of the attention mechanism.

The visual encoder processes image patches using self-attention mechanism to capture global visual features. Specifically, the query \( Q_{\text{img}} \), key \( K_{\text{img}} \), and value \( V_{\text{img}} \) are obtained via learned linear projections \( W_Q, W_K, W_V \), respectively. The attention weights are computed as:
\begin{equation}
A_{\text{Vision Encoder}} = \text{Softmax} \left( \frac{Q_{\text{img}} K_{\text{img}}^T}{\sqrt{d}} \right)
\end{equation}
In contrast, the LLM decoder utilizes causal self-attention, which is essential for autoregressive tasks such as text generation. The input to the decoder typically includes tokens from the system prompt, image tokens, textual input, and previously generated output tokens. The corresponding query \( Q_{\text{input}} \), key \( K_{\text{input}} \), and value \( V_{\text{input}} \) are computed analogously. However, causal attention is enforced by applying a lower triangular causal mask \( M \), ensuring each token attends only to itself and preceding tokens:
\begin{equation}
A_{\text{LLM Decoder}} = \text{Softmax} \left( \frac{Q_{\text{input}} K_{\text{input}}^T + M}{\sqrt{d}} \right)
\end{equation}

\subsection{Does Image Token Pruning Necessarily Lead to Visual Information Loss?}

In general, image tokens carry aligned visual-language information that is essential for tasks such as image captioning and visual question answering. Intuitively, reducing the number of these tokens, regardless of the method or extent, could increase the risk of hallucinations or degraded outputs. However, our experimental findings suggest that this assumption does not always hold.

To investigate this, we employed the FasterVLM pruning strategy, which is based on pruning image tokens and the [CLS] token before the LLM processing. We evaluated model performance across a range of pruning ratios, from 10\% up to 90\% and 95\%. Surprisingly, we observed that  shallow pruning—removing a small percentage of tokens—often led to improved performance. We attribute this to the removal of redundant or overly fine-grained visual details through image-to-image pruning, which helps filter out noise and reduces unnecessary computational burden. These redundant tokens, if retained, may distract the language model and ultimately impair its reasoning ability. 

However, as the pruning ratio increases, a significant drop in performance emerges. This decline is attributed to the loss of essential visual cues that are critical for downstream understanding. These results highlight the importance of a balanced pruning strategy—one that carefully removes irrelevant information while preserving task-relevant visual content.

\begin{figure}[h!]
    \centering
    \begin{subfigure}[b]{0.46\textwidth}
        \includegraphics[width=\linewidth]{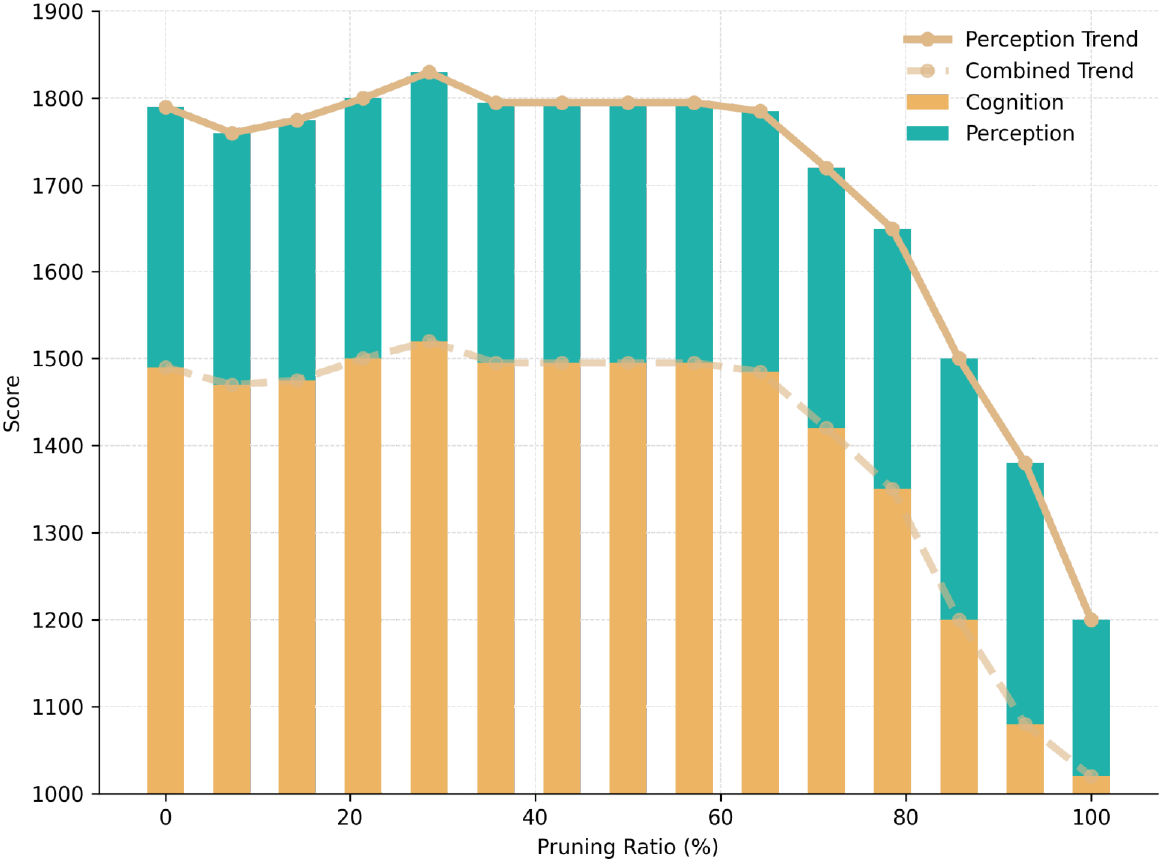}
        \caption{}   
        \vspace{-2mm}
    \end{subfigure}
    \hfill
    \begin{subfigure}[b]{0.52\textwidth}
        \includegraphics[width=\linewidth]{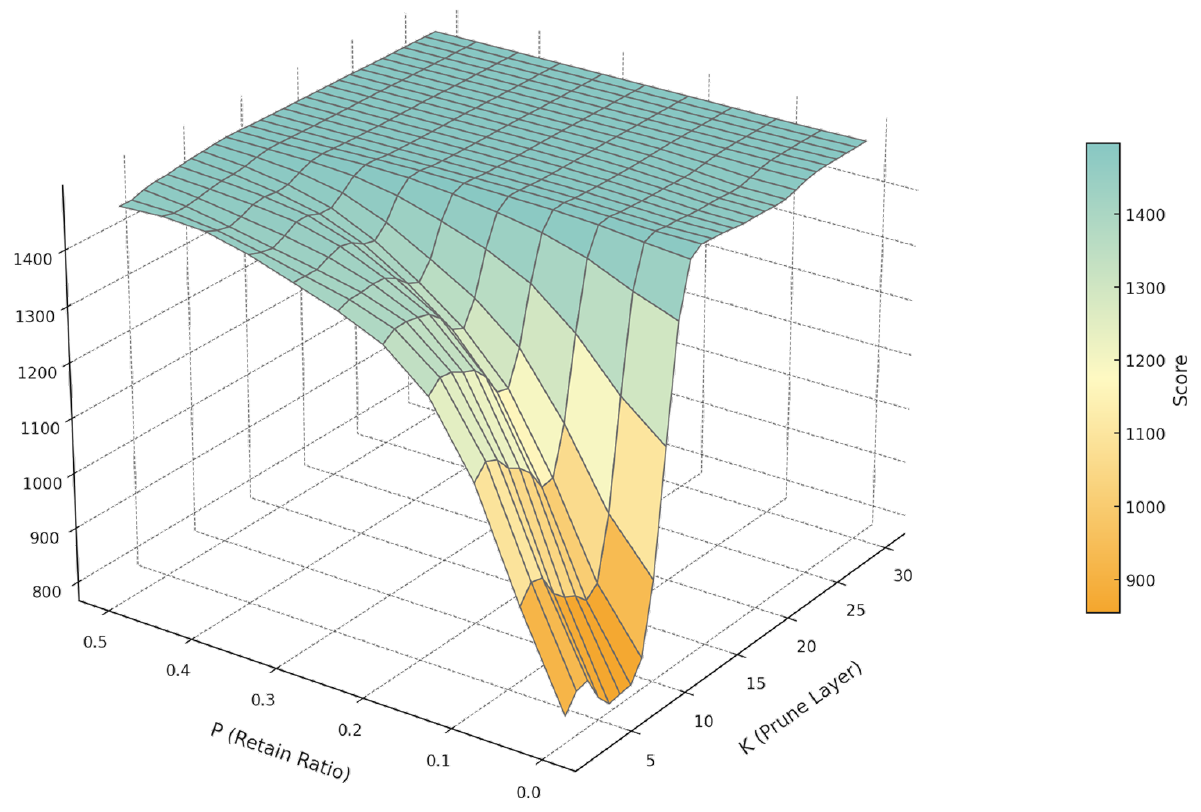}
        \caption{}
        \vspace{-2mm}
    \end{subfigure}
    \caption{(a) Model performance under different pruning ratio of image tokens before the LLM decoder. (b) Impact of pruning layer and image tokens retain ratio on model performance.}
    \label{fig:motivation}
\end{figure}

\subsection{Which Layer to Prune?}

FastV first reveals the inefficiency of visual tokens in LVLMs, particularly noting that  visual tokens receive significantly less attention than textual tokens in the deeper decoder layers. Motivated by this phenomenon, it prunes visual tokens based with low attention scores after the second decoder layer to reduce computational cost and improve inference efficiency.

However, while FastV successfully identifies the inefficiency of visual tokens and suggests pruning based on text-visual attention, our experiments results demonstrate that pruning visual tokens too early in the model (in the earlier layers) does not yield the best performance. Instead, pruning at deeper layers, particularly where the image and text tokens receive sufficient attention from the model, proves to be more effective. By pruning image tokens in intermediate layers, we can significantly reduce the low-attention tokens that contribute misleading information in deeper LLM layers, thus preventing the model from producing misleading results.
As shown in Figure \ref{fig:motivation}, when pruning visual tokens after deeper layers of the LLM, the model retains more of the useful visual information while reducing redundant tokens. This approach not only decreases inference cost but also improves model performance, especially in cases where the reduction ratio is high.

\section{Method}
\begin{figure}[h]
  \centering
  \includegraphics[width=1\linewidth]{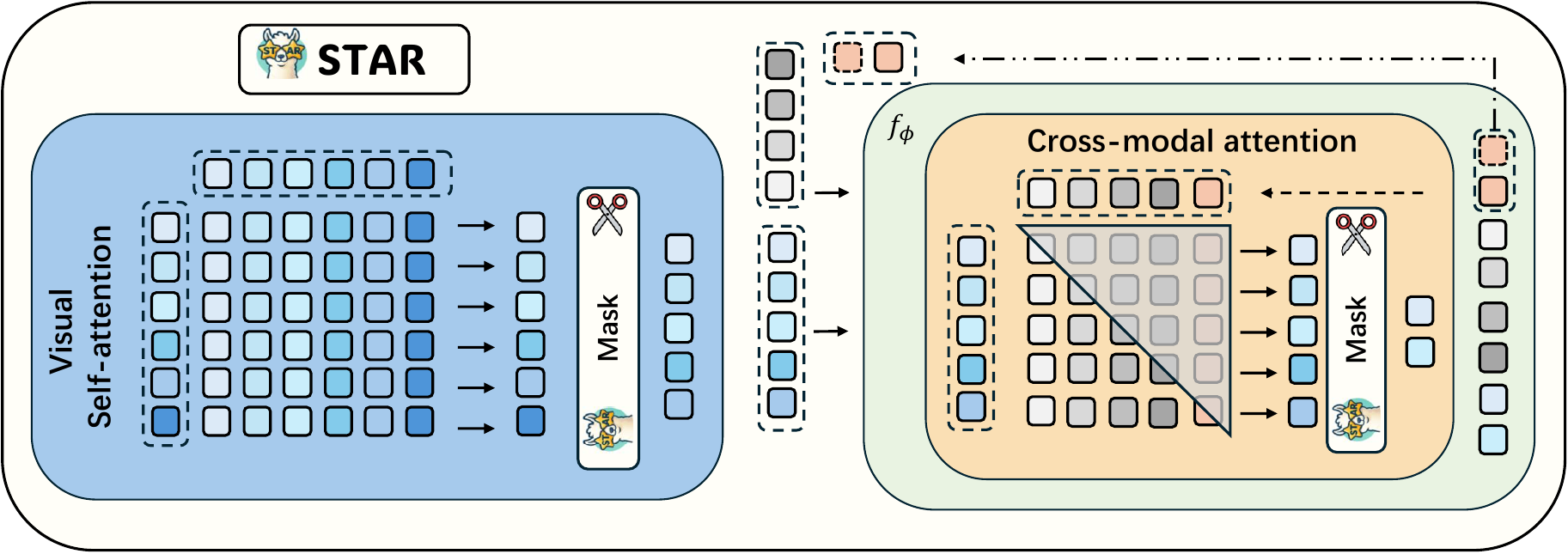}

  \caption{
\textbf{The implementation of STAR mechanism.} In the first stage, an importance score indicator is computed to guide the removal of low-level redundant features. In the second stage, we use the cross-modal importance indicator to select important visual tokens, which is strongly aligned to input query and generated response.}
  \label{fig:method}
\end{figure}

We introduce \textbf{STAR}, a two-stage token-pruning framework designed to accelerate inference in Large Vision–Language Models (LVLMs). In Section \ref{sec:visual_self_attention}, we compute a visual-only importance score $r_i$ by averaging self-attention weights of each image token and prune the least informative tokens using a dynamic thresholding strategy. Section \ref{sec:cross_modal_attention} further refines the token set by evaluating cross-modal attention between the remaining visual tokens and the concatenated text query–response stream at a selected decoder layer, removing tokens with low vision–language relevance. In section \ref{sec:Theoretical}, we also detail the thresholding criteria, the multi-modal projection step that prepares the pruned tokens for the LLM decoder, and conclude with a theoretical analysis of the resulting reduction in FLOPs. An overview of the STAR framework is illustrated in Figure~\ref{fig:method}.

\subsection{Visual Self-Attention}\label{sec:visual_self_attention}
The visual encoder produces a set of token embeddings $H_v \in \mathbb{R}^{L_v \times d}$
and the corresponding self‐attention map $A \in \mathbb{R}^{L_v \times L_v}$. To estimate the importance of each visual token, we compute an importance score indicator $\tilde{r} = [r_1, \dots, r_{L_v}]^\top \in \mathbb{R}^{L_v}$ by averaging the rows of \(A\):
$$
A = \mathrm{Softmax}\!\bigl(\tfrac{H_v H_v^\top}{\sqrt{d}}\bigr), \qquad
r_i = \frac{1}{L_v} \sum_{j=1}^{L_v} A_{ij}, \quad
i = 1,\dots,L_v.
$$
Here, \(A_{ij}\) denotes the attention weight from token \(i\) to token \(j\), and \(r_i\) reflects the overall importance of \(i\)-th visual token. Given a user-specified token reduction ratio \( R \in (0, 1) \), we determine a dynamic threshold \( \tau \) to retain only the top \( (1 - R) \cdot L_v \) tokens:
\begin{equation}
  \tau
  \;=\;
  \min\Bigl\{\tau' \;\bigl|\;
      \bigl|\{i:r_i\ge\tau'\}\bigr|
      \le (1-R)\,L_v
  \Bigr\},
  \label{eq:ta_tau}
\end{equation}
We then prune the least informative tokens by retaining only those with \( r_i \ge \tau \):
$$
Z_v' = \{\,z_i \in Z_v \mid r_i \ge \tau\}\,.
$$
The remaining visual tokens \( Z_v' \) are passed through a multi-modal projection layer \( g \), resulting in visual features \( H_v = g(Z_v') \), which are concatenated with the textual instruction embeddings \( H_q \) to form the decoder input:
\begin{equation}
  X_{\mathrm{input}}
  = \operatorname{Concat}\bigl(H_v,\,H_q\bigr)
  = \operatorname{Concat}\bigl(g(Z_v'),\,H_q\bigr).
  \label{eq:ta_concat}
\end{equation}
This stage performs early token reduction before feeding tokens into the LLM decoder, effectively discarding low-level redundant features and significantly reducing computational overhead.

\subsection{Cross-Modal Attention}\label{sec:cross_modal_attention}

Let \(H_v \in \mathbb{R}^{L_v \times d}\) denote the visual embeddings and let $H_q \in \mathbb{R}^{L_t \times d}$
and $H_{\text{resp}} \in \mathbb{R}^{L_o \times d}$ represent the embeddings of the textual query and generated response tokens, respectively. We concatenate these two textual streams to form  
\(
\widetilde{H}_q = [\,H_q \,;\, H_{\text{resp}}\,] \in \mathbb{R}^{(L_t+L_o)\times d}.
\)
Using \(\widetilde{H}_q\), the cross-modal attention map at decoder layer~\(K\) is
\begin{equation}
C_K \;=\;
\operatorname{Softmax}\!\Bigl(
  \frac{ H_v \,\widetilde{H}_q^{\!\top} }{ \sqrt{d} }
\Bigr)
\;\in\;
\mathbb{R}^{L_v \times (L_t+L_o)},\quad
\label{eq:cross_attn_resp}
\end{equation}
To compute a second-stage importance indicator, we take the average attention score across all textual positions for each visual token. Specifically, for token \( i \), we define:

$$
r_i = \frac{1}{L_t + L_o} \sum_{j=1}^{L_t + L_o} C_K[i,j],
\quad i = 1,\dots,L_v,
$$
where \( C_K[i,j] \) represents the attention from visual token \( i \) to the \( j \)-th token in the combined query–response sequence. By stacking all scores into a vector \( \tilde{r} = [r_1, \dots, r_{L_v}]^\top \in \mathbb{R}^{L_v} \), we obtain the cross-modal importance indicator. To meet a target pruning ratio \( P \in (0,1) \) at layer \( K \), we discard the \( P \cdot L_v \) tokens with the lowest scores in \( \tilde{r} \), retaining the top \( (1 - P) \cdot L_v \) most relevant visual tokens. 

This stage ensures that only visual tokens with strong alignment to both the input query and the model-generated response are propagated to subsequent layers, further enhancing computational efficiency while preserving cross-modal semantic integrity.

\subsection{Theoretical Computation Reduction}\label{sec:Theoretical}

Following \cite{zhang2024sparsevlm}, we adopt the single–FLOP‐per‐MAC rule (\textsc{mac}=1) and focus on the two dominant components in a Transformer decoder layer: multi‐head self‐attention and the feed‐forward network (FFN). For sequence length \(L\) and hidden width \(D\) (FFN inner size \(=D\)), the baseline FLOPs per layer is
\[
F_{\mathrm{base}}
=6\,L\,D^{2} \;+\;2\,L^{2}D,
\]
where the term \(6LD^{2}\) accounts for the six linear projections \(\{\mathbf Q,\mathbf K,\mathbf V,\mathbf O,W_{1},W_{2}\}\) and \(2L^{2}D\) corresponds to the matrix multiplications involved in attention computation. When pruning \(N_i\) visual tokens at decoder layer \(i\) (without recycling, i.e., \(C_i = 0\)), the exact reduction in FLOPs is:
\[
\Delta_i
=6\,N_{i}\,D^{2} \;+\;2\,N_{i}^{2}\,D,
\]
We consider pruning across \(\Omega\) decoder layers, with the process split at a designated pivot layer \(K\) into two stages. In Stage 1 (layers \(1\) to \(K\)), a fixed text-agnostic pruning ratio \(R\) is applied. In Stage 2 (layers \(K+1\) to \(\Omega\)), a more adaptive, cross-modal pruning ratio \(P\) is used. Given \(L_v^0\) original visual tokens, the number of tokens pruned per layer is:
\[
N_i
=
\begin{cases}
R\,L_v^0, & 1 \le i \le K,\\[4pt]
P\,L_v^0, & K+1 \le i \le \Omega.
\end{cases}
\]

The cumulative FLOPs reduction in each stage is:
\[
\Delta_{\mathrm{stage1}}
=K\bigl[6\,R\,L_v^0\,D^{2}+2\,R^{2}(L_v^0)^{2}D\bigr],
\qquad
\Delta_{\mathrm{stage2}}
=(\Omega-K)\bigl[6\,P\,L_v^0\,D^{2}+2\,P^{2}(L_v^0)^{2}D\bigr].
\]

Adding these yields the total forward‐pass FLOPs saved:
\[
\Delta_{\mathrm{total}}
=
\underbrace{K\bigl[6D^{2}L_v^0R+2D(L_v^0)^{2}R^{2}\bigr]}_{\textbf{Stage 1}}
\;+\;
\underbrace{(\Omega-K)\bigl[6D^{2}L_v^0P+2D(L_v^0)^{2}P^{2}\bigr]}_{\textbf{Stage 2}}.
\]
In our design, we enforce \(R < P\), reflecting a deliberate choice to apply conservative, text-independent pruning before visual tokens enter the language model decoder. This preserves low-level visual richness in early layers while allowing more aggressive, semantically guided pruning in deeper decoder layers where vision–language fusion is more mature.

\section{Experiment}
\subsection{Experimental Setup}

\paragraph{Datasets}
In order to verify that the application of STAR mechanism does not degrade model performance, we conduct extensive experiments on eight widely-used multimodal VQA benchmarks, including VQAv2 \cite{goyal2017makingVQAV2}, GQA \cite{hudson2019gqa}, VizWiz \cite{gurari2018Vizwiz}, ScienceQAIMG \cite{lu2022learnScienceQA}, TextVQA \cite{singh2019towardsTextVQA}, POPE \cite{li2023evaluatingPOPE}, MME \cite{fu2023mme}, and MM-Vet \cite{yu2023mmvet}. These experiments on various datasets provide a comprehensive assessment of the effectiveness and generality of STAR.

\paragraph{Models and Method}
We apply STAR to various representative LVLMs, including LLaVA-1.5 \cite{liu2023visual} with 7b and 13b parameters, and LLaVA-NeXT \cite{liu2024llavanext} with 7b parameters for high-resolution image inputs. For comparison, we consider three efficient token pruning methods: FastV \cite{chen2024imagefastv}, FasterVLM \cite{zhang2024clsFastervlm}, and SparseVLM \cite{zhang2024sparsevlm}. Specifically, FastV identifies visual redundancy and removes low-attention tokens early in the LLM layers. FasterVLM ranks visual tokens using [CLS]-to-patch attention scores from the visual encoder and prunes low-salience tokens before passing them to the LLM. SparseVLM, on the other hand, progressively sparsifies visual inputs based on cross-modal attention to reduce noise.

\subsection{Main Results}

We first apply STAR to the LLaVA-1.5 and conduct a comprehensive comparison with existing token reduction methods. Table \ref{tab11} presents the performance of LLaVA-1.5-7b with different numbers of remaining visual tokens (from 576 to 29). As SparseVLM enforces a minimum of 29 visual tokens via its recycling mechanism, we omit SparseVLM from the most aggressive pruning setting (29 tokens) in our comparison.

\begin{table*}[h!]
  \centering
  \caption{Performance on general vision-language tasks with LLaVA-1.5-7B (R=0.1, K=14).}
  \label{tab11}
  \scriptsize
  \setlength{\tabcolsep}{7pt}

  \begin{tabular}{
      c
      l
      *{7}{c}
  }
  \toprule
  \textbf{Rem.\ Visual Tokens} & \textbf{Method} 
    & \textbf{VQAv2} & \textbf{GQA} & \textbf{VizWiz} & \textbf{TextVQA}
    & \textbf{SQA-IMG} & \textbf{MME} & \textbf{POPE} \\
  \midrule
  576
    & Baseline
    & 78.52 & 61.93 & 50.06 & 58.25 & 69.56 & 1491.55 & 79.80 \\
  \midrule
  \multirow{4}{*}{288}
    & FastV             & 77.66 & 60.10 & 50.58 & 58.27 & \bfseries 69.01 & 1485.34 & \bfseries 80.70 \\
    & SparseVLM         & 77.02 & 59.33 & \bfseries 51.32 & 57.53 & 68.82 & 1418.12 & 79.40 \\
    & FasterVLM         & 77.86 & 60.61 & 50.40 & 57.87 & 68.37 & 1440.17 & 78.80 \\
    & \textbf{STAR (Ours)} & \bfseries 78.40 & \bfseries 61.77 & 50.07 & \bfseries 57.80
                        & 68.96 & \bfseries 1506.09 & 79.60 \\
  \midrule
  \multirow{4}{*}{115}
    & FastV             & 72.41 & 55.21 & 51.78 & 56.60 & 68.77 & 1357.92 & 77.80 \\
    & SparseVLM         & 74.63 & 56.47 & 51.76 & 56.35 & 70.25 & 1349.98 & 77.90 \\
    & FasterVLM         & 75.37 & 57.51 & \bfseries 52.17 & 56.65 & 68.42 & 1398.94 & 77.20 \\
    & \textbf{STAR (Ours)} & \bfseries 78.18 & \bfseries 61.71 & 49.32 & \bfseries 57.96
                        & \bfseries 69.01 & \bfseries 1497.14 & \bfseries 79.40 \\
  \midrule
  \multirow{4}{*}{58}
    & FastV             & 65.16 & 51.12 & 51.73 & 54.64 & \bfseries 69.81 & 1176.03 & 69.30 \\
    & SparseVLM         & 63.85 & 49.23 & 47.52 & 49.91 & 69.81 & 1293.00 & 77.40 \\
    & FasterVLM         & 71.93 & 54.89 & \bfseries 50.05 & 55.27 & 68.91 & 1314.73 & 76.80 \\
    & \textbf{STAR (Ours)} & \bfseries 77.68 & \bfseries 61.04 & 48.67 & \bfseries 57.79
                        & 69.16 & \bfseries 1505.23 & \bfseries 79.50 \\
  \midrule
  \multirow{3}{*}{29}
    & FastV             & 55.40 & 45.66 & 48.92 & 51.23 & \bfseries 70.15 & 949.89  & 57.20 \\
    & FasterVLM         & 66.76 & 51.54 & \bfseries 52.71 & 53.11 & 69.51 & 1183.68 & 72.90 \\
    & \textbf{STAR (Ours)} & \bfseries 76.45 & \bfseries 59.51 & 47.24 & \bfseries 56.79
                        & 68.86 & \bfseries 1512.35 & \bfseries 79.80 \\
  \bottomrule
  \end{tabular}
\end{table*}

The results show that all single-stage token-reduction methods noticeably degrade model performance, indicating that solely relying on single-stage pruning is insufficient to identify and preserve informative tokens. In contrast, STAR significantly mitigates this performance decline. For example, on the MME dataset, with 29 remaining tokens, STAR improves performance by +20.8 points over the baseline and surpasses FastV by +562.46 points. Moreover, on the comprehensive dataset VQAv2 and GQA, STAR consistently achieves the best results across various settings of remaining visual tokens, demonstrating its robustness and strong generalization capability.

Our design results in only minor degradation in end-to-end performance, in a specific intermediate layer $K$. In contrast, other methods failed to fully utilize sustained cross‐modal attention, leading to performance degradations. These observations highlight that two-stage pruning framework appears to be more effective in preserving performance.

We further evaluate STAR method on the larger LLaVA-1.5-13B model, comparing it with FastV \cite{chen2024imagefastv} and FasterVLM \cite{zhang2024clsFastervlm} across representative benchmarks—TextVQA \cite{singh2019towardsTextVQA}, SQA-IMG \cite{lu2022learnScienceQA}, MME \cite{fu2023mme} and POPE \cite{li2023evaluatingPOPE}.

As shown in Table~\ref{tab:llava-1.5-13b}, STAR remains highly robust as the decoder depth increases. When the number of visual tokens is reduced from $576$ to $29$, STAR exhibits almost no performance degradation on the POPE and TextVQA benchmarks, with scores dropping by only 0.5\% and 0.9\% respectively.

As summarized in Table \ref{tab:llava-next-7b}, we evaluate STAR on LLaVA-Next-7B, which represents a more challenging tasks which processing much higher-resolution inputs.

As image resolution rises, FasterVLM discards too much visual information, causing sharp accuracy drops on all four benchmarks. FastV is more resilient under heavy pruning, yet its efficiency gains plateau once only a few tokens remain. In contrast, STAR achieves the best trade-off: it sustains near-baseline accuracy with negligible degradation while providing substantial inference speed-ups, achieving the strongest results on the MME benchmark.


\begin{table*}[t!]
  \centering
  \captionsetup{justification=centering}
  \tiny
  \setlength{\tabcolsep}{3pt}

  \caption{Performance comparison on four vision–language benchmarks
           for LLaVA-1.5-13B (left) and LLaVA-Next-7B (right)}
  \label{tab22}

  \begin{subtable}[t]{0.48\textwidth}
    \centering
    \subcaption{LLaVA-1.5-13B (R = 0.1, K = 14)}
    \label{tab:llava-1.5-13b}
    \resizebox{\linewidth}{!}{%
      \begin{tabular}{c l c c c c}
        \toprule
        \textbf{Rem.\ Tok.} & \textbf{Method}&
        \textbf{TextVQA} & \textbf{SQA-IMG} & \textbf{MME} & \textbf{POPE} \\
        \midrule
        576 & Baseline& 61.19 & 72.78 & 1505.03 & 79.1 \\
        \midrule
        \multirow{3}{*}{288}
          & FastV                & 60.87 & 73.13 & \textbf{1508.71} & 79.9 \\
          & FasterVLM            & 59.99 & \textbf{73.62} & 1499.51 & 77.6 \\
          & \textbf{STAR (Ours)} & \textbf{61.10} & 72.93 & 1488.46 & \textbf{79.9} \\
        \midrule
        \multirow{3}{*}{115}
          & FastV                & 59.08 & \textbf{74.37} & 1464.49 & 80.7 \\
          & FasterVLM            & 58.55 & 73.97 & 1425.32 & 76.6 \\
          & \textbf{STAR (Ours)} & \textbf{60.75} & 72.83 & \textbf{1488.11} & \textbf{79.4} \\
        \midrule
        \multirow{3}{*}{58}
          & FastV                & 55.60 & 72.24 & 1351.96 & 77.2 \\
          & FasterVLM            & 57.35 & \textbf{73.72} & 1344.19 & 75.6 \\
          & \textbf{STAR (Ours)} & \textbf{60.23} & 72.68 & \textbf{1495.69} & \textbf{78.9} \\
        \midrule
        \multirow{3}{*}{29}
          & FastV                & 51.85 & \textbf{73.18} & 1144.73 & 70.1 \\
          & FasterVLM            & 54.84 & 72.78 & 1218.02 & 72.5 \\
          & \textbf{STAR (Ours)} & \textbf{59.69} & 72.78 & \textbf{1479.83} & \textbf{78.2} \\
        \bottomrule
      \end{tabular}%
    }
  \end{subtable}
  \hfill
  \begin{subtable}[t]{0.48\textwidth}
    \centering
    \subcaption{LLaVA-Next-7B (R = 0.1, K = 14)}
    \label{tab:llava-next-7b}
    \resizebox{\linewidth}{!}{%
      \begin{tabular}{c l c c c c}
        \toprule
        \textbf{Rem.\ Tok.} & \textbf{Method}&
        \textbf{TextVQA} & \textbf{SQA-IMG} & \textbf{MME} & \textbf{POPE} \\
        \midrule
        576 & Baseline& 59.67 & 69.56 & 1513.21 & 85.06 \\
        \midrule
        \multirow{3}{*}{288}
          & FastV                & 61.50 & 69.01 & \textbf{1516.14}& \textbf{85.53}\\
          & FasterVLM            & 59.75 & 69.66 & 1514.90 & 85.13 \\
          & \textbf{STAR (Ours)} & 59.52 & \textbf{69.91} & 1503.41 & 84.70 \\
        \midrule
        \multirow{3}{*}{115}
          & FastV                & \textbf{60.84}& 68.77 & \textbf{1504.00}& 85.56\\
          & FasterVLM            & 59.24 & 68.96 & 1477.14 & \textbf{85.66}\\
          & \textbf{STAR (Ours)} & 59.31 & \textbf{70.05} & 1503.55 & 84.60 \\
        \midrule
        \multirow{3}{*}{58}
          & FastV                & \textbf{60.56}& 69.81 & 1481.62 & \textbf{85.40}\\
          & FasterVLM            & 57.60 & 68.07 & 1375.13 & 83.73 \\
          & \textbf{STAR (Ours)} & 59.18 & \textbf{69.96} & \textbf{1504.80}& 84.73 \\
        \midrule
        \multirow{3}{*}{29}
          & FastV                & \textbf{59.67}& \textbf{70.10}& 1472.29 & \textbf{85.06}\\
          & FasterVLM            & 55.92 & 68.82 & 1220.06 & 79.13 \\
          & \textbf{STAR (Ours)} & 58.97 & 69.81& \textbf{1505.41}& 84.80 \\
        \bottomrule
      \end{tabular}%
    }
  \end{subtable}
\end{table*}


\begin{table*}[h!]
  \centering
  \scriptsize
  \setlength{\tabcolsep}{6pt}
  \caption{Inference efficiency comparison between FastV, FasterVLM, and STAR (Ours) with LLaVA-1.5-7B on MME benchmark.}
  \label{tab:inference_efficiency}

  \begin{tabular}{c l c c c c c}
    \toprule
    \textbf{Rem.\ Visual Tokens} & \textbf{Method} & \textbf{FLOPs (T)} & \textbf{Avg.\ FLOPs (T)}
                          & \textbf{GPU Mem (GiB)} & \textbf{Latency (ms)} & \textbf{Throughput (it/s)} \\
    \midrule
    576 & LLaVA-1.5-7B & 21353.56 & 8.99 & 15.24 & 135.58 & 7.38 \\
    \midrule
    \multirow{4}{*}{288}
      & FastV              & 13471.31 & 5.67 & \textbf{13.79} & 130.11 & 7.69 \\
      & FasterVLM          & \textbf{11967.71} & \textbf{5.04} & 14.99 & \textbf{103.35} & \textbf{9.68} \\
      & SparseVLM          & 12116.37 & 5.10 & 18.31 & 142.74 & 7.01 \\
      & \textbf{STAR (Ours)} & 15223.38 & 6.41 & 14.47 & 134.68 & 7.42 \\
    \midrule
    \multirow{4}{*}{58}
      & FastV              & 13471.31 & 5.67 & \textbf{13.79} & 129.92 & 7.70 \\
      & FasterVLM          &  \textbf{4620.35} & \textbf{1.95} & 15.04 & \textbf{ 73.06} & \textbf{13.69} \\
      & SparseVLM          &  5167.20 & 2.18 & 18.28 & 116.87 &  8.56 \\
      & \textbf{STAR (Ours)} & 11434.42 & 4.82 & 14.35 & 123.89 &  8.07 \\
    \midrule
    \multirow{3}{*}{29}
      & FastV              & 13471.31 & 5.67 & \textbf{13.79} & 129.74 &  7.71 \\
      & FasterVLM          &  \textbf{3703.29} & \textbf{1.56} & 15.04 &  \textbf{69.16} & \textbf{14.46} \\
      & \textbf{STAR (Ours)} & 10965.06 & 4.62 & 14.52 & 121.80 &  8.21 \\
    \bottomrule
  \end{tabular}
\end{table*}

\subsection{Computational Efficiency}
\label{sec:computational_efficiency}

Table~\ref{tab:inference_efficiency} compares the inference efficiency of our STAR method against the vanilla LLaVA-1.5-7B, FastV~\cite{chen2024imagefastv} and FasterVLM~\cite{zhang2024clsFastervlm} on the MME benchmark, measured on a single NVIDIA RTX 4090 24 GB GPU with identical text-prompt lengths and single-image inputs. Running LLaVA-1.5-7B with all 576 visual tokens requires a total of 21 353.56 TFLOPs (8.99 TFLOPs per image on average), occupies 15.24 GiB of GPU memory, and incurs 135.58 ms of latency (7.38 it/s). 

FasterVLM removes a large number of visual tokens before the LLM, thereby most effectively reducing computational overhead: it cuts total compute by 43.9 \% (to 11 967.71 TFLOPs) and latency to 103.35 ms (9.68 it/s), though this comes at the expense of noticeable performance degradation.

By comparison, FastV masks visual tokens during the decoder layers, reducing total compute by 36.9 \% (to 13 471.31 TFLOPs) and latency by 4.0 \% (to 130.11 ms, 7.69 it/s). However, this masking-based approach is constrained by throughput and latency limitations, resulting in slower overall inference.

STAR, by contrast, cuts total compute by \textbf{28.7} \% (to 15 223.38 TFLOPs), saves 0.77 GiB of memory (14.47 GiB) with only a 0.7 \% latency increase (to 134.68 ms), and maintains 7.42 it/s. Under aggressive pruning to 58 tokens, STAR requires just 11 434.42 TFLOPs (15.1 \% less than FastV), runs in 123.89 ms (4.6 \% faster) and sustains 8.07 it/s. Even at the extreme 29-token budget, STAR maintains 10 965.06 TFLOPs (–18.6 \% vs. FastV), 121.80 ms latency (–6.1 \%), and 8.21 it/s—demonstrating substantial reductions in both TFLOPs and CUDA time.

\begin{figure}[t]
\vspace{-3mm}
  \centering
  \includegraphics[width=1\linewidth]{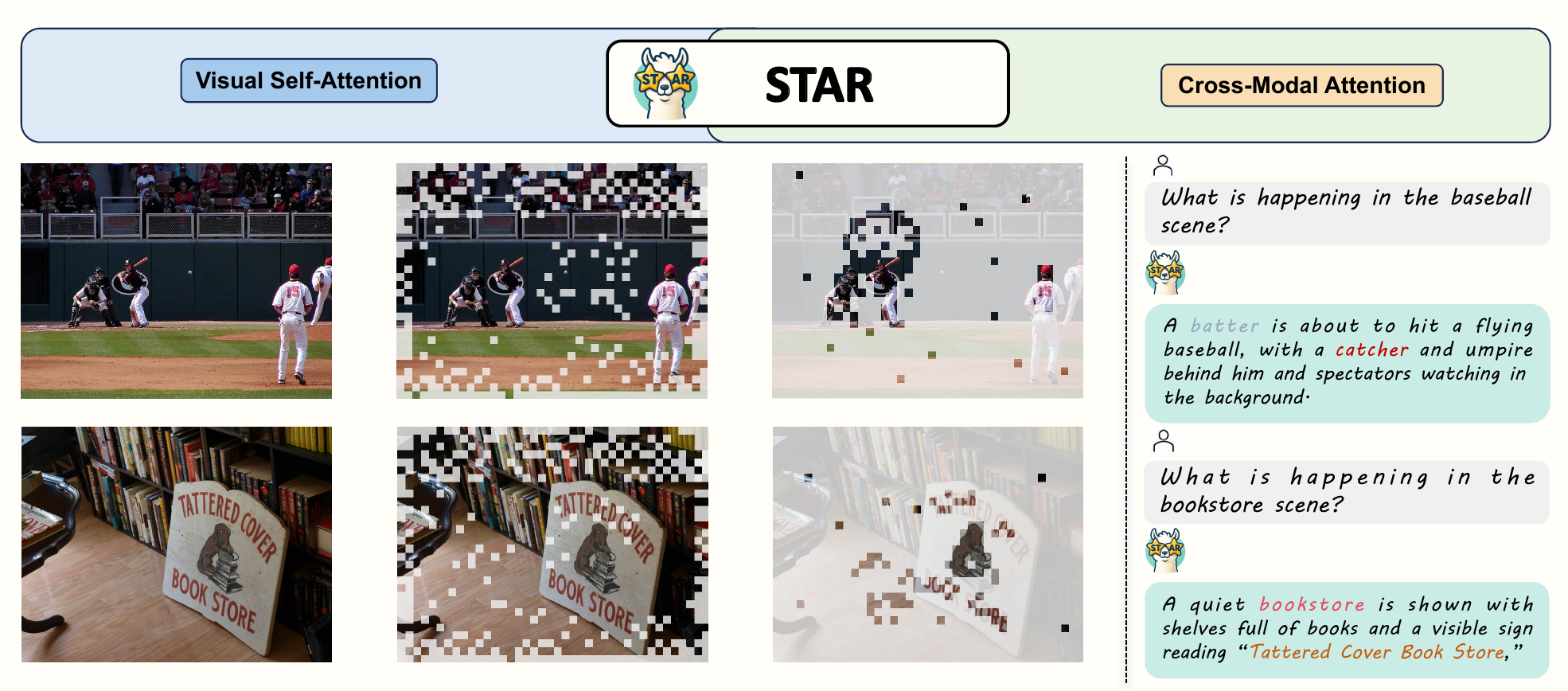}
  \caption{Visualization of the STAR method: Examples generated using the Vicuna-7B model}
  
  \label{fig:Visualization of Star}
\vspace{-3mm}
\end{figure}

\subsection{Visualization of STAR}

As shown in Figure \ref{fig:Visualization of Star}, we visualize the remaining tokens of STAR on various VQA questions. The image results are visualized after token pruning while before LLM, and at a specific intermediate layer in LLM. In the early stage, low-level redundant features are pruned, while retaining features with representative information.
In the later stage, STAR prunes task-irrelevant tokens based on the cross-attention mechanism. The model systematically reduces less relevant image information while retaining key tokens closely tied to the question. 
The visualization reveals that STAR, although discarding some overall image details, effectively retains essential visual tokens. These preserved tokens encapsulate the features necessary for answering the question, focusing on more relevant visual regions through their interaction with the question.

\section{Conclusion}

In this paper, we present \textbf{STAR}, a plug-and-play, training-free token reduction framework for LVLMs. By applying attention-guided reduction in two stages—first using visual self-attention to drop redundant low-level tokens, then using cross-modal attention to discard task-irrelevant ones—STAR delivers substantial inference speed-ups with negligible impact on downstream accuracy. Experimental results demonstrate that STAR consistently outperforms existing token pruning methods across a wide range of vision–language benchmarks such as MME, POPE and VQAV2. This illustrates particularly strong robustness of STAR under extreme pruning ratios.  Furthermore, STAR demonstrates stable generalization on large-scale models like LLaVA-1.5-13B and high-resolution models such as LLaVA-Next-7B. From the perspective of inference efficiency, STAR achieves substantial acceleration with minimal performance degradation.

\bibliographystyle{unsrtnat}
\bibliography{refs}

\end{document}